\documentclass[runningheads]{llncs}
\usepackage[T1]{fontenc}
\usepackage{graphicx}
\begin{document}
\title{Inching Towards Automated Understanding of the Meaning of Art: An Application to Computational Analysis of Mondrian's Artwork}
%
%
\author{Alex Doboli\inst{1}\and
Mahan Agha Zahedi\inst{1}\and
 Niloofar Gholamrezaei\inst{2}}
\authorrunning{F. Author et al.}
%
\institute{Stony Brook University, Department of Electrical and Computer Engineering, Stony Brook NY 11794-2350, USA\\
\email{\{alex.doboli,mahan.zahedi\}@stonybrook.edu} \and
School of Art, Texas Tech University, Lubbock TX, USA\\
\email{niloofar.gholamrezaei@ttu.edu}}
\maketitle              
\begin{abstract}
Deep Neural Networks (DNNs) have been successfully used in classifying digital images but have been less succesful in classifying images with meanings that are not linear combinations of their visualized features, like images of artwork. Moreover, it is unknown what additional features must be included into DNNs, so that they can possibly classify using features beyond visually displayed features, like color, size, and form. Non-displayed features are important in abstract representations, reasoning, and understanding ambiguous expressions, which are arguably topics less studied by current AI methods. This paper attempts to identify capabilities that are related to semantic processing, a current limitation of DNNs. The proposed methodology identifies the missing capabilities by comparing the process of understanding Mondrian's paintings with the process of understanding electronic circuit designs, another creative problem solving instance. The compared entities are cognitive architectures that attempt to loosely mimic cognitive activities. The paper offers a detailed presentation of the characteristics of the architectural components, like goals, concepts, ideas, rules, procedures, beliefs, expectations, and outcomes. To explain the usefulness of the methodology, the paper discusses a new, three-step computational method to distinguish Mondrian's paintings from other artwork. The method includes in a backward order the cognitive architecture's components that operate only with the characteristics of the available data.

\keywords{classification of art  \and computational methods \and cognitive architecture.}
\end{abstract}

\section {Introduction}

Classifying items based on their defining characteristics as well as identifying these characteristics has been a major research topic in Machine Learning (ML)~\cite{Goodfellow2016}. Driven by applications in computer vision and text processing, like text summarization and translation, numerous ML algorithms have been devised including both procedural methods as well as data modeling techniques. The latter methods compute the parameters of  parameterized models to minimize the error between training sets and model predictions. Deep Neural Networks (DNNs) pertain to this approach, including Convolutional Neural Networks (CNNs) a popular type of DNNs.

CNNs have been successful in classifying digital images~\cite{LeCun2010}. However, recent work explored the CNN's capability to classify images with meanings that are not linear combinations of their visualized features~\cite{Mahan2022}. This work showed that CNNs cannot correctly distinguish artwork presenting complex, hard-to-grasp non-exhibited properties (NEXP), even though they perform well for art objects described mainly by exhibited properties (EXP). EXP are visual features, like form, color, scene, and NEXP represent meaning, like an artist's intention and an observer's perception~\cite{Mahan2022}. For example, paintings from the Renaissance period display a rich set of EXPs, and abstract paintings, because of their abstraction, include NEXPs that are hard to learn. Even though there has been a hope that DNNs capability to be universal approximators will somehow support the picking-up of EXP combinations that describe well NEXPs too, experiments showed that such descriptions are not learned with current CNNs~\cite{Mahan2022}. 

It is unclear what features are missing from present CNNs, so that they can effectively identify and use NEXPs in classification of artwork. In spite of artificially generated art~\cite{Bentley2002}, e.g., using Genetic Algorithms and other evolutionary algorithms, it is arguable if the produced outputs are similar to the artwork created by humans and analyzed by domains, like aesthetics~\cite{Baxandall1985}. It has been argued that art expresses human experiences~\cite{Baxandall1985,Mondrian1945,Mondrian1995}, which is obviously not the case with generated artwork. The meaning of human experiences relates to human intention and understanding, which depend on numerous historical, economic and social factors~\cite{Baxandall1985,Mondrian1945}. For example, intentional historical theory argues that an artist had an intention to create art of a certain kind~\cite{Levinson1993}. Then, this kind must be inferred (understood) by the viewer~\cite{Bloom1996}. It is possible that dissimilar artwork (including dissimilar EXPs) still belongs to the same kind~\cite{Bloom1996}. Hence, it is important to understand how knowledge about meaning (semantics) is used by humans during the process of creating (intention) and understanding (perception) art. 

We believe that the significance of creating computational methods towards understanding how NEXPs (e.g., meaning) operate in art is well beyond creating immediate applications, like automatically generating museum inventories, artwork explanations, and interactive avatars to guide visitors. NEXPs are tightly connected to abstract representations, reasoning, and understanding of abstract and ambiguous expressions, which are different in nature than processing and learning EXPs (i.e. using features on form and structure for classification), the focus of current AI methods. Abstract reasoning and understanding ambiguous expressions are critical in human problem solving by individuals and teams of individuals. 

Starting from the observation that current DNN approaches are arguably insufficient to tackle semantic abstractions~\cite{Mahan2022}, this work focused on identifying the capabilities that must be added to DNNs to improve their capabilities of semantic processing. The used methodology identifies the missing features by comparing the process of understanding artwork with the process of understanding and creating electronic circuits, a well-known domain of creative problem solving. Our previous work proposed a cognitive architecture, a computational structure loosely based on human cognitive reasoning, to automatically synthesize electronic circuits~\cite{Li2018}. We leveraged this work to propose a cognitive architecture meant to understand the abstract paintings of Piet Mondrian. The comparison of the two processes referred to two semantic layers: The first layer has eight elements: goals, concepts, ideas, rules, procedures, beliefs, expectations, and outcomes. These elements are part of the cognitive process during problem solving and can be further linked to more detailed cognitive activities, like memory, concept learning, representation and combination, affect, insight, and so on. The eight elements are discussed and compared for the process of understanding and performing circuit design and the process of understanding Mondrian's paintings. The second layer describes the solving process and includes five elements: the nature of the problem, knowledge representation in the memory, the attention and prediction subsystem, the reasoning subsystem, and knowledge updating. The comparison of the elements of the two semantic layers indicates what capabilities must be modified or added to a cognitive architecture geared towards understanding Mondrian's paintings as compared to the architecture in~\cite{Li2018}. The functional requirements and evaluation metrics for these capabilities can be then stated. 

We argue that this work is a step towards connecting a computational approach to understanding abstract paintings by Mondrian to cognitive activities, even though it is not a connection to the neural activity of the brain, as in neuroaesthetics~\cite{Baxandall1985,Onians2007,Zeki1993}. Subsequent work can attempt to relate the cognitive activities to DNN processing, similar to~\cite{Eliasmith2013}. We believe that Mondrian's work is appropriate for this goal: EXPs are less important as it is visually simple (e.g., uses vertical ad horizontal lines and surfaces colored with fundamental colors, e.g., red, blue, yellow, and white), but complex in terms of its meaning defined using NEXPs. Note that other work discusses problem solving in domains, like physics and mathematics~\cite{Polya1957,Swanson1990,Wang2010}, but does not propose equivalent algorithmic methods even though they discuss common-sense strategies to solve problems.               

The paper has the following structure. Section~2 presents related work. Section~3 offers an overview of the model used to identify the requirements of computational methods to classify using meaning. Section~4 focuses on a case study that compares the computational needs for electronic circuit design and Mondrian's paintings. A discussion follows in Section~5. Conclusions end the paper.        

\section {Related Work}

Modern ML methods, like CNNs, have been proposed to automatically analyze artwork, including activities, like style recognition, classification, and generation~\cite{Lecoutre2017,Spratt2015,Noord2014,Zhao2021}. As large training sets are often hard to assemble for art, the traditional approach pre-trains a CNN using large databases of images, e.g., ImageNet, and then retrains only the output and intermediate layers using art images~\cite{Lecoutre2017,Roberto2020,Noord2014,Zhao2021}. 

Style recognition finds the artistic style of artwork using mainly visual attributes, like color and texture~\cite{Lecoutre2017,Roberto2020,Noord2014,Zhao2021}. Seven different CNN models were tested for three art datasets to classify genres (e.g., landscapes, portraits, abstract paintings, etc.), styles (i.e. Renaissance, Baroque, Impressionism, Cubism, Symbolism, etc.), and artists~\cite{Zhao2021}. The method uses mostly color. It achieves for some styles an accuracy similar to human experts. However, other styles are hard to recognize, like Post Impressionism and Impressionism, Abstract Expressionism and Art Informel, or Mannerism and Baroque~\cite{Lecoutre2017}. CNN are also suggested to recognize non-traditional art styles, like Outsider Art style~\cite{Roberto2020}.

CNNs are used to identify the author from a group of artists by learning the visual features of the artist's work~\cite{Noord2014}. The method utilizes features, like texture, color, edges, and empty areas~\cite{Noord2014}. However, it is necessary to also use higher-level features, like localized regions or semantic features, e.g., scene content and composition~\cite{Saleh2016}.

Work on uncovering semantic information about artwork intends to understand the content of art objects, like the orientation of an object, the objects in a scene, and the central figures of an object~\cite{Gonzalez2018,Lelievre2021,Simonyan2015}. Object orientation, e.g., deciding if a painting is correctly displayed, uses simple, local cues, image statistics, and explicit rules~\cite{Lelievre2021,Liu2017,Zafar2018}. The methods have been reported to be as effective as human interpretation for some painting styles~\cite{Lelievre2021}. They perform better in portrait paintings than in abstract art. The distinguishing features among classes include localized parts of large objects, low intra-class variability of the parts, and specific semantic parts, such as wheels for cars and windows for buses. Generative Adversarial Networks (GANs) are proposed for hierarchical scene understanding~\cite{Yang2021}. Early layers are likely to learn physical positions, like the spatial layout and configuration, the intermediate layers detect categorical objects, and the latter layers focus on scene attributes and color schemes. 

\section{Model Components}

Creating works of art, like paintings, can be seen as the process of solving open-ended problems~\cite{Baxandall1985}, similar to open-ended problem-solving in engineering, like imagining and devising new functional capabilities that are beyond those of the current solutions. Both represent instances of human creative activities. Subsection~3.1 compares conceptually the two types of problem-solving endeavors. Next, we presented the elements used in the comparison.  

Culture includes the goals, concepts, ideas, rules, procedures, beliefs and outcomes of a certain population, and the expectations of their outcomes~\cite{Baxandall1985}. (1)~Goals are high-level objectives (problems) in response to needs. For example, it has been mentioned that one of the goals for art is to produce pleasure~\cite{Onians2007,Zeki1993} or to make a donor proud~\cite{Baxandall1985}. (2)~Concepts are the building blocks of knowledge. They are characterized by common, defining features and by features that distinguish them from other concepts~\cite{Doboli2014,Doboli2015}. (3)~Ideas are sets of related concepts that serve/produce a certain purpose, e.g., enable a certain situation or create a certain output. (4)~Rules indicate the way of relating concepts to each other, and (5)~procedures are sequences of rules that produce a desired outcome. (6)~Beliefs are ideas and rules that are somewhat constant (invariant) within a certain culture. (7)~An outcome is an expression (materialization) of cultural components in an object of art as well as the degree to which the new expression differs or improves of previous similar expressions. (8)~An expectation correlates with the priority associated to a cultural component, and subsequently to an emotion. Multiple expectations can exist within a population, which results in multiple priorities and emotions. Expectations can be common to a population or can be different depending on the subjects experiences~\cite{Baxandall1985,Onians2007}. 

{\bf Example}: We referred to Baxandall's discussion of the painting ``Baptism of Christ'' by Piero della Francesci~\cite{Baxandall1985}. Baxandall presents the cultural environment in which the painting was produced. (1)~It includes the painter's goal in response to a customer's request, which refers to a place of display (e.g., an altar piece), a topic (i.e. Christ's baptism), and a specific artist to create the artwork. As Baxandall explains in~\cite{Baxandall1985}, the painting was meant to have three functions: ``to narrate scripture clearly, to arouse appropriate feeling about the narrated matter, and to impress that matter on the memory'' (page 106). (2)~Concepts include all items that form the ``language'' of a culture, like Christ, angels, baptism, water, and so on, and their associated meanings, including symbolisms. (3)~Ideas are a set of concepts, in this painting, the idea that Christ's baptism by a human indicates his humility~\cite{Baxandall1985}. The meaning of the idea might result through inference based on its concepts, through the using of analogies and metaphors, and so on. (4)~Rules include the application of mathematical principles to painting, such as rules related to perspective, proportions, and Euclidean analysis of forms~\cite{Baxandall1985}. Onians offers an interesting presentation of the evolution of the rules embedded into artwork, starting from empirical rules defined by ancient Greeks and Romans, and followed by rules based on geometry, the physiology of the eye and brain, psychology, and neuroscience~\cite{Onians2007}. (5)~Procedures include the established templates and routines for devising a painting, like selection of colors, its design, and composition~\cite{Baxandall1985}. Procedures also include the way of assigning purpose and intention to an art object, such as a certain way of interpreting its meaning through causal inference~\cite{Baxandall1985}. (6)~Beliefs refer to certain widely accepted meanings and facts, like the interpretation of Scripture. (7)~The outcome is the actual painting in this example. (8)~As Baxandall explains, expectations include the following attributes: ``clear, moving, memorable, sacramental, and creditable image of the subject'' (page 106 in \cite{Baxandall1985}). Expectations can also refer to the characteristics of the physical place of display, like its size, shape, etc. .    

Arguably, the state-of-the-art in technology is the equivalent of culture, as it includes the goals pursued by a certain technological domain, its laws, theories and models, procedures, and designs. Expectations are based on the functionality and performance of the previous designs. Similarly to culture, expectations define a priority of the components defining the state-of-the-art, and an emotion originating in the satisfaction of the expectations through the actual designs. Multiple expectations are possible as community members can have different predictions on how previous designs will shape future outcomes.         

{\bf Example}: We referred to the electronic circuit designs discussed in~\cite{Doboli2015,Jiao2015,Li2018,Umbarkar2014}. The state-of-the-art, which describes the context in which the circuits were discussed, includes the following elements that are analogous to the elements discussed for the previous example. (1)~The goal was to design circuits with open-ended functions, like the problem description referred to using state-of-the-art technology for embedded systems and a set of building blocks (BBs), like sensors and actuators, to improve campus life~\cite{Umbarkar2014}. However, once the desired functionality is identified, the associated performance requirements become defined too. Performance requirements are numerical values that must meet minimum (maximum) limits or are minimized (maximized) as part of the design process. (2)~Concepts are the BBs used in design, like MOSFET transistors, resistors, capacitances, and subcircuits. Similar to concepts in art, they can have multiple meanings, e.g., as a MOSFET transistor can have multiple functions (behaviors) depending on its region of operation, however, the semantics of these meanings is much simpler than in art. Also, their meanings can be described in a precise, formal way using mathematics or logic. (3)~Ideas represent a set of BBs with a precise meaning, like subcircuits built out of MOSFETs. (4)~Rules indicate how BBs should be connected to each other in a design, such as the rules that describe the connections of the four MOSFET transistor terminals. (5)~Procedures represent the steps to be pursued to solve a design problem, like the steps to create a circuit structure (schematics) and the steps to size the MOSFET transistor. (6)~Beliefs refer to the paradigms considered to be valid for the considered state-of-the-art, such as its functional and performance capabilities as well as its advantages and disadvantages compared to other alternative state-of-the-art in technology. (7)~Outcomes is the set of all circuit designs that have been created for the given state-of-the-art. (8)~Expectations refer to the degree to which a designed electronic circuit meets its functional and performance requirements.        

\subsection{Semantics of Creative Activities in Painting and Electronic Circuit Design}



This section compares the semantic elements of the creative processes in painting and electronic circuit design. 

\subsubsection{(1) Goals:} High-level goals are expressed through the set of concrete problems that must be solved. These problems are characterized by four variables: topic, relationship to previous problems, physical constraints, and authorship. 

The topics of paintings might encompass a broad range that goes from a well-defined set of ideas and symbolisms, i.e. the meaning that results from Scripture for Christian religious paintings, to a broad range of emotions and meanings that attempt to establish a dialog between the artwork and viewers, like in abstract painting~\cite{Baxandall1985}. Therefore, while the end goal of a painting is to produce pleasure to the viewer~\cite{Baxandall1985,Onians2007}, it can achieve it by using a broad range of meanings, on one end based on a well-defined meaning ({\em closed semantic space}) and at the other end based on primary cues that originate a sensation of pleasure in the brain ({\em physiological space of the brain}), like color~\cite{Onians2007,Zeki1993}. 

In engineering, problem framing is an important design step and mainly refers to defining the precise functionality of the solution and its expected performance~\cite{Gao2004}. Problem-framing can co-evolve with the solving process~\cite{Baxandall1985,Dorst2001}. New problems are often defined with respect to previously developed designs ({\em incremental design}). While there are relationships to previous work, there is arguably more flexibility in defining the topic of a new art object, as there are no concrete functional or performance requirements other than being original and improving the well-being and pleasure of customers~\cite{Baxandall1985} ({\em physiological space of the brain}). As opposed to art where topic identification is usually decided by the customer and/or artist~\cite{Baxandall1985}, polls are used in circuit design to identify appealing functions, or the actual opportunities might become evident only after a number of attempts, such as in mobile computing.   However, as opposed to art, after the function of a design is well defined, the degree to which the goal is attained can be characterized through well-defined metrics on cost, accuracy, speed, energy and power consumption, reliability, and so on~\cite{Gielen2000}. 

The place of display of a painting can be well defined, like it was for ``Baptism of Christ'', or might be initially undefined, as in the case of a painting an artist creates on his / her own. However, in both cases, the physical dimensions of the painting are set before the artist commences the painting. Similarly, the physical attributes of an electronic design, i.e. weight and size, are decided after its functionality and envisioned way of use are decided. As opposed to art, where physical constraints are fixed ({\em constraint satisfaction problem}), e.g., the painting must entirely occupy the assigned 2D surface, in electronic design, the goal might be to minimize the size and weight ({\em minimization problem}). 

Finally, the author of an artwork is well defined ({\em fixed}) while the authors of an electronic circuit design do not indicate a certain person but rather refer to a group of persons that possess a required skill set ({\em constraint satisfaction}).      

\subsubsection{(2) Concepts:} Concepts are collections of items that share common features and are distinguishable from other concepts~\cite{Doboli2015,Ferent2013,Markman1997}. Studying concepts has been a main topic in cognitive psychology~\cite{Cohen1984}. There are various concept models, like using prototypes to represent concepts~\cite{Rouder2006}, class membership models based on primitives and defining conditions~\cite{Ashby2005,Cohen1984,Ferent2013,Rouder2006}, class membership models using graded membership~\cite{Ward2002}, and concept typicality and vagueness~\cite{Murphy2005,Osherson1997}. Concept formation and learning discusses topics, like concept alignment~\cite{Lassaline1998}, concept similarity~\cite{Goldstone1991} and dissimilarity~\cite{Ranjan}, perceptual symbols~\cite{Barsalou1999}, natural categories~\cite{Rosch1976,Ross1999}, ad-hoc categories~\cite{Barsalou1983}, goal-based concepts~\cite{Barsalou1991}, incremental formation~\cite{Gennari1989}, and complex concept formation~\cite{Murphy1988}. Work also studied the difference between physical concepts and its image in the memory~\cite{Ward2002}, situated simulation~\cite{Barsalou2003} and thematic relations~\cite{Lin2001}, categorization adaptation~\cite{Anderson1991}, and concept typicality~\cite{Hampton1988}. Next, the differences between concepts in paintings and electronic circuit design were discussed with respect to their possible meanings, descriptions, and meaning understanding. 

In~\cite{Baxandall1985}, Baxandall states that ``In systems like classical mythology and Christian theology, matured and elaborated over centuries, almost anything can signify something - trees, rivers, the various colors, groups of twelve, seven, three and even one; many things can signify various things'' (page 132). For example, he indicates seven different meanings for Baptism, and the various meanings of a dove, plants, forms, and colors ({\em meaning transfer from another space})~\cite{Baxandall1985}. The many meanings of the concepts originate a very large of possible meanings for the composition of forms (concepts) in a painting ({\em size of the semantic space}). Moreover, different concepts can have similar meanings, therefore there is significant redundancy in the semantic space ({\em redundancy of the semantic space}). In contrast, the concepts (e.g., BBs) in an electronic design can have multiple meanings, but the number of possible meanings is small. For example, a MOSFET transistor can act as a switch that is on or off or as an amplifying element. However, these meanings (behaviors) are few and in general restricted to a certain BB. Moreover, the meaning of the concepts in paintings is tightly integrated with the other concepts to achieve unity and balance, including requirements expressed in the laws of aesthetics~\cite{Dickie1997} and Gestalt theory~\cite{Arnheim1974} ({\em integration}). A concept's meaning must be consistent with that of the entire composition, and new meanings can result in this process. BBs can be tightly coupled to each other, like in electronic designs of small size, like analog circuits, but modularization through local and hierarchical coupling of BBs was proposed to manage the complexity of large size design, thus, to achieve scalability ({\em scalability requirement}). The meaning of concepts in art can continuously change over time and from one culture to another due to dialog between the artist, participants ({\em members of the same culture as the artist}), and observers ({\em members of a different culture})~\cite{Baxandall1985} ({\em dynamics of meanings}). A circuit design has a precise meaning that rarely changes over time ({\em unless new applications are discovered for it}), however, there can be a translation process from one state-of-the-art to another when designs are migrated across different fabrication technologies~\cite{Gielen2000}. The translation is then always complete, while certain painting features might be difficult to map to a new culture, like the concept of ``commensurazione'' as explained in~\cite{Baxandall1985}.  

In design, the meaning of a BB is often specified using formal descriptions, like closed-form mathematical expressions (e.g., differential equations), logic formulas (i.e. first or higher-order logic), and executable / simulatable specifications in a programming language (like VHDL-AMS)~\cite{Gielen2000}. These descriptions are often grounded in laws of physics. When such descriptions are not available, the BB meaning is expressed by enumerating the behavior in the main cases, like the corner points of a MOSFET device. In general, it can be argued that methods have been devised to produce tractable descriptions of behavior, so that they cover as much as possible of the possible behavior of a BB. Unexpected BB behaviors are unwanted as they usual end-up in failures, and additional design features might be incorporated into an electronic circuit to avoid such situations ({\em maximize the deterministic, fully known semantics} of BBs). In contrast, the features of the concepts in painting are rarely fully and well-defined, as enabling the inference of multiple meanings of a painting's concepts and composition is a main goal of art~\cite{Onians2007,Zeki1993}. Having new, unusual concept features, like a purple tree or a strange posture of an angel, can express new meanings, and are part of the creative process in art. Also, the level of detailing of a concept (and vice versa its level of abstract representation) can be important in painting for conveying a certain message, like emphasizing a certain aspect. In contrast, BBs in a complete engineering design are fully specified, even though partial descriptions can be used during the previous drafting stages. The physical representation of concepts in paintings can be subjected to geometrical rules, like placement of objects and perspective, which is similar to devising the structure of a circuit and sizing the BBs in electronic design. However, such rules are not imposed, e.g., in modern art, while they are a strong requirement in circuit design. 

Finally, the process of understanding the meaning of a concept in painting involves inference and using symbols, analogies, and metaphors. Insight is gained in the process. Baxandall explains that this process is similar to formulating hypotheses, and then validating them~\cite{Baxandall1985}. The process is a sequence of steps, in which the meaning of a composition produces a hypothesis that is applied top-down to understand the meaning of concepts. Any identified inconsistencies serve then bottom-up as cues to modify the hypothesis and re-execute the process. The end result is a story that explains a painting. There are similarities with understanding the meaning of BBs in electronic designs, as humans might formulate hypotheses about the purpose (e.g., function) of unknown BBs, and then verifying the hypotheses, followed by readjusting a hypothesis if needed and repeating the process. However, it can be argued that the formulated hypotheses are simpler due to the significantly less ambiguous meaning of BBs than concepts in painting. This likely impacts the utilized strategies to find valid hypotheses.     

\subsubsection{(3) Ideas:} Ideas are sets of related concepts assembled to satisfy a certain purpose, like to enable something or to create an output or consequence. Hence, an idea has an associated meaning. The purpose and meaning can be different for different cultures, including changes over time. For example, the idea of Christ's baptism expresses the cleansing of sins in Christian religions with all the associated consequences, like the ``unbaptised are damned'' (in~\cite{Baxandall1985}, page 123). Or, the idea that a painting meant to be an altarpiece has a precise function. Similarly, in electronic circuit design, collections of related BBs (concepts) form a design with a precise purpose, like functionality and performance. Idea representation and organization has been studied in psychology~\cite{Anderson1980,Barsalou1992,Petrou2010}. 

The discussed comparison of ideas in painting and electronic circuit design considered the following dimensions: the type of ideas as defined by the nature of their purpose, meaning, origin, characteristics, grouping, and evolution (change). Moreover, ideas can be explicit or implicit. Explicit ideas are stated and communicated to others, hence become part of the shared culture or state-of-the-art in circuit design. For example, the laws of proportions and perspective were ideas that were stated and shared by medieval painters. Similarly, textbook descriptions of subcircuits represent ideas on how to obtain a certain functionality and performance. Implicit ideas refer to situations in which the idea produces a well-defined purpose without that this purpose was consciously intended by the artist or designer ({\em emergence}). Also, a painter might prefer a certain color or  organization of a composition. Similarly, designers might prefer a specific subcircuit, even though other possibly superior alternatives exist. 

Regarding their type, ideas in art can serve different purposes, like inquiry, hypothesizing, explanation (including causality), critique, constraint, generalization, detailing, expectations, purposing, intention, confirmation, support, reinforcement, messaging (propagation), expression of emotions, and so on. For example, ideas can express how a certain intention is produced through the selected colors and organization. Depending on their purpose, e.g., problem solving or knowledge communication, ideas in circuit design can have the same types, with less emphasis on expression of emotions.  

It can be argued that meaning can be defined as a black box model, if it only explains the purpose of an idea (what), or a white box model, if it explains how the purpose is achieved (how). For example, Gombrich argued the importance of forming a visual image of an idea~\cite{Gombrich1982b}. The meaning of ideas result through different processes in painting and circuit design. In art, the meaning is tightly dependent of the context, such as the time when an artwork was created and interpreted, and the perspective~\cite{Gombrich1984,Onians2007}. Different meanings can result depending on the viewer's perspective. New insight is likely to result. Therefore, ambiguity is arguably a desirable features, as it supports novel, constructive and creative reinterpretations of an art object. Moreover, physiological and psychological reaction can be important, such as body reactions, unconscious reactions to shapes and colors, and empathy~\cite{Gombrich1982,Onians2007}. Therefore, it can be argued that the meaning of an artwork is not self-contained, but itself results from the subjective interpretation of the observer embedded in his/her culture. Instead, the meaning of ideas in circuit design is well-defined, mostly self-contained, and less dependent on possible interpretations. While multiple meanings can exist, alternatives are usually few, known, and precisely defined.  

Ideas also differ by the process used to assign them a meaning, i.e. to understand them. Ideas in paintings are understood through processes that involve hypothesis formulation and verification, using symbols, analogies and metaphors, abstraction, utilizing exaggerations, paradoxes and contradictions, satire, allusions, logic inference, identifying associations, and so on~\cite{Baxandall1985}. Hence, it is important to formulate, focus, and pursue multiple meanings through cognitive activities, like separation, classification, and understanding the whole before focusing on the parts (top-down reasoning)~\cite{Onians2007}. The importance of sequential understanding has been also argued, especially for modern art~\cite{Baxandall1985}. Note that there is a physiological biasing of the process due to the brain's built-in priority in focusing on forms, like face and eyes~\cite{Onians2007}. The importance of idea organization in the memory, as well as the differences between experts and amateurs have been also explained in the literature~\cite{Onians2007}.  

Idea characteristics include the following attributes: Precision refers to the degrees of ambiguity, such as in the case of metaphors, as well as if an idea has a quantitative or qualitative evaluation. Ideas can describe different degrees of abstraction, and can have different levels of rigidity depending on their flexibility to support changes and to be combined (related) with other ideas. Their invariance describes the degree to which ideas remain unchanged as a result of the idea combination process, e.g., their meaning remains the same in spite their change. They can have different degrees of validity, like ideas which are always true (truisms), are valid under certain circumstances, and are always false. They have a certain organization and structure, including a hierarchical structure. Ideas can have different social attributes too, cogency (degree of authority), visibility, importance, and impact for a culture. 

An idea is usually part of a larger group of ideas, therefore it has characteristics with respect to the ideas of the entire group. Idea similarity describes the similarity of two ideas with respect to their concepts, structure, or meaning. Hence, similarity expresses the degree to which ideas are aligned with each other, including situations in which an idea is a transformed or evolved descendant of another idea. Ideas described by different sets can have similar meanings (synonyms). Alternatively, distinct ideas are described by their degree of differentiation. The capability to compare two ideas with respect to a metric, such as their utility, supports the definition of an idea's quality. Familiarity is the frequency with which an idea has been repeated within a culture. Ideas can be organized (structured) in an ontology, such as hierarchies of clusters of similar ideas~\cite{Doboli2015}. Groups of ideas can be described by patterns, which express conditions valid for all ideas in the group. Ideas can be characterized by their consistency, e.g., the degree to which their meanings are logically not conflicting (contradicting) with each other, continuity which is the possibility of understanding them as an evolving sequence, and integrality which is the degree to which the set specifies a unitary, complete system (ensemble). Ideas can be described by a degree of unexpectedness (including oddity), like whether they can be predicted based on the context or other ideas, and degree of redundancy, such as their meaning is articulated by other ideas too.     

Ideas can be learned through experience, from others, or obtained through insight. Ideas can change over time and can be common, opposite, different, and partially different for the members of the same culture or of different cultures. The degree of idea similarity can depend on conditions of the context. Baxandall explains that there is a continuity over time between ideas in art, such as an idea can be related to previous ideas~\cite{Baxandall1985}. Also, ideas in art can be continuously reinterpreted for any new time period~\cite{Baxandall1985}.     

\subsubsection{(4) Rules:} Rules present a way to connect concepts, ideas, or other rules with each other. The latter two kinds are called metarules. Concept combinations have been studied in cognitive psychology as a mechanism to relate separate concepts~\cite{Mumford1997}. There are two types of combinations: property-based and relation-based combinations~\cite{Wisniewski1998}. Property-based combinations transfer features from one concept to another while the interpretation is plausible~\cite{Osherson1982}. Relation-based combinations connect two nouns through modifiers that relates to causes, structure, purpose, and location~\cite{Gagne2000}. The parameters that influence the selection of different relation types has been also studied, like cueing, stimuli sequencing, memory~\cite{Gagne2002}.

Rules are described in our model by the following elements that are detailed next: conditions for applying the rule, including constraints, a rule's structure and elements, the expected goals and real outcomes of applying a rule, the characteristics of a rule, its interpretation, and its origin and gradient (change). 

The conditions for applying a rule describe the situations (conditions) under which a rule becomes available to be used and then selected to be used. In general, models distinguish between making a rule available (cuing / activating it)~\cite{Gagne2000} and selecting (deciding) the rule from the set of available rules~\cite{Gagne2000,Gagne2002}. Constraints can cause the activation of a rule, like a certain structure of a painting's physical display (like the specific place of display of an altar piece in a church) which imposes rules on the structuring of the composition~\cite{Baxandall1985}. Rules are also activated by a certain pursued goal and established beliefs about how the goal can be achieved. For example, art was dominated for some time by the belief that paintings must be nature-accurate images, which triggered the need to apply rules on proportion and perspective, so that 3D images were accurately presented on 2D surfaces~\cite{Baxandall1985}. The using of rules is also decided by the meaning associated in certain contexts and cultures, like the connection between the importance of an object and the centrality of its representation in a composition. Universal rules are always true, even though they might not be always selected, such as in situation when the author wants to communicate a paradox or an oddity. 

A rule's structure and elements indicate its constitution. Rules can connect concepts, features of concepts, and ideas to a certain outcome, or connect rules into higher level structures, like hierarchical structures. For example, rules can express the structure of a composition or design. When higher-level structures are created, the connection can use all the components of the lower elements (lossless case) or only some (loss case) while possibly adding new elements that are not present in the lower structures (extension case). 

The expected goals and real outcomes of applying a rule can be global, if it refers to an entire image, or local, if it relates to certain parts and details. Depending on their structural activity, rules can serve to decompose and aggregate an entire, to produce associations between concepts and ideas, to compare and differentiate, to generalize instances and to de-generalize (instantiate) abstractions. With respective to their cognitive goal, rules can serve as part of producing inquiry, hypothesis, analysis, discussion, pairing and comparison (including alignment, similarity, separation / difference, and prominence / superiority), explanation, necessity, evaluation, insight, articulation and persuasion, reinforcement (memorizing), support, predictions, expectations, achieving a certain sentiment, impression, and emotion, and so on. Depending on the preciseness of their expectations and goals, rules can introduce different levels of ambiguity (including metaphors, symbols, analogies, and allusions), backward references to old goals, and new meanings and reinterpretations of an existing rule. Specific rules can communicate truth. Rules can be used to communicate repetition, patterns, or movement, such as placing opposite colors at the opposing ends of horizontal and vertical lines~\cite{Deicher1999}. Rules can also suggest correspondences between different images, between an image in nature and its corresponding artistic image, between outputs and goals, between attention and action.  

The characteristics of a rule include whether a rule is explicit or implicit, externalized or internalized. Explicit rules consciously identify the involved elements (i.e. concepts, ideas or other rules), while the precise nature of their connection might (i.e. through mathematical expressions) or might not be defined (e.g., through qualitative or approximate expressions). Implicit rules are executed without the awareness of the user. Externalized rules are those described in a communication media, like formal rules, speech, image, and so on. Internalized rules do not have such a description. Rules can be deterministic or stochastic, which is a consequence of their activation and selection processes. Self-contained rules are fully expressed only based on their description. Rules can be also approximate (if they describe reality within a certain error range), precise (if they express the desired relation without unnecessary or redundant information), structuring (if there is an organization of large rules, i.e. using hierarchical structures), robust (if they are supported by a large set of real-world situations suggesting a certain organization of the experience), flexible (if they can be changed into other rules), invariant over conditions, including time (if the rule remains valid for a broad set of changing conditions, i.e. contexts and cultures), and relatable to other rules (if there is a sequence of rules that establishes the connection between two rules, i.e. deductive reasoning, inference, contradiction, or exploring alternatives). 

The interpretation of a rule indicates the meaning of a rule, as described by its structure, concepts, effects, and conditions of application. As a rule can have multiple meanings and ambiguity, depending on the degree of consistency (validity) of its meanings, rules are needed to assign and update rigor, address unexpected features, strangeness / peculiarities, oddities, seeming, paradoxes, and their connection to attention. The meaning of a rule can change depending on the broader context and the user's subjective experience and perspective. Rules can also link material, natural elements to internal, subjective representations, such as in using lines, positions, and colors the achieve harmony, balance, and ultimately beauty~\cite{Mondrian1995}, or communicating the feeling of grandiosity through a rule that decides the viewing position of the viewer with respect to the painting~\cite{Mondrian1995}. Rules can also decide the target of the viewer's attention during observation, such as through large surfaces of the same color or by placing objects in the central part of a painting. Rules can produce a global viewing of an entire painting (overview) or guide viewing towards local details, such as through the crowding and dimensions of the emerging forms and surfaces. For example, many, small patches of colors can suggest an interpretation focus on details, while few, large surfaces encourage a global view of the entire painting.              

Rules can originate in math and sciences or in various social conventions. For example, it is argued that the rules used in art were grounded over time in geometry, physiology of the eye and brain, psychology, and neuroscience. Alternatively, rules have been grounded in beliefs, like rules on the origin of harmony and beauty~\cite{Mondrian1995}, or communicating a certain moral or ethical message~\cite{Baxandall1985}. Rules can be learned from others, like mentors, or learned through own experience, like using certain brush strokes to convey a certain message or the way of constructing compositions to achieve the desired visual harmony and balance of the composition~\cite{Mondrian1995}. Rules can be also set by the client of an art object. Another aspect is the degree to which new rules are embraced by the larger community, the degree of propagation, and the degree to which a rule changes over time as a result of personal and collective experience and depending on the conditions and context it is applied, the rule's flexibility. An interesting interplay arises between the goals of an outcome and rule modification (evolution). For example, the goal of creating nature-accurate paintings supported the continuous search for rules that would accurately represent a 3D image on a 2D surface. Note that this problem requires finding rules to project from a higher dimension space into a space of lower dimension with minimum information loss.

\subsubsection{(5) Procedures:} Procedures represent problem-solving sequence of steps, where each step applies rules to concepts and ideas. Problem solving has been a main research topic in psychology~~\cite{Gigerenzer1999,Jonassen2011,Simon1971,Simon1979,Stenning2008}. The studied topics include the nature (e.g., clauses like what (declarative), how (procedural), etc.) and organization of memory structures (including differences between experts and novices~\cite{Chi1981,Cross2004,Ho2001,Schvaneveldt1985,Swanson1990} and memory cuing~\cite{Chi1981,Heller1982}), learning~\cite{Langley2005,Sun2005}, such as concept formation~\cite{Chi1981}, new rule induction~\cite{Chi1981} and priority formation~\cite{Swanson1990}, dual-process reasoning~\cite{Osman2004}, solving heuristics and their motivations~\cite{Chen1999,Gigerenzer1999}, problem solving processes through problem decomposition based on categorization using similarity with previous problems~\cite{Chi1981,Liikkanen2009,Schvaneveldt1985}, solution synthesizing through an ordered, sequenced tackling~\cite{Liikkanen2009} of categorized schemata, templates and selected knowledge snippets~\cite{Gobet2002,Lawson2004,Simon1979,Swanson1990,Yamauchi2000} and solution verification through redundant perspectives~\cite{Heller1982}, and understanding, including insight~\cite{Bowden2005,Schilling2005,Suwa2001,Thagard2011} and knowledge restructuring~\cite{Little2006,Vosniadou1987}.  

The comparison of the procedures used in problem solving in art and circuit design refers to the following elements: the devising of the step sequence to solve a problem, connection to previous and future problem solving instances, and learning and getting insight while conducting the procedure. 

There is abundant literature on problem solving procedures in mathematics, science, and engineering~\cite{Polya1957,Wang2010}. These approaches are geared towards solving well-defined and ill-defined problems with known requirements and constraints. For example, the steps suggested in~\cite{Polya1957} require first identifying the problem variables and structure, followed by relating it to similar, previously solved problems. Problem solving using analogies is often used too~\cite{Vosniadou1989}. If the problem cannot be solved, the process should attempt solving a simpler yet related problem, created either by decomposing the original problem (divide-and-conquer) or by simplifying the problem while keeping its main variables and unknowns (problem approximation). Depending on how problems are simplified and then their solutions integrated into the final solutions, the process uses various nuances of concept combinations, in which property combinations decide the features of the concepts (BBs) used in the solution~\cite{Osherson1982}, and relation-based combinations decide how concept functions are integrated together~\cite{Wisniewski1998}. Using methods grounded in logic, such as inference, is an example of relation-based combination. However, other ways to achieve relation-based combinations are possible too, as explained in~\cite{Jiao2015,Li2018}. Verifying the correctness of the solution ends the process. Relation-based combinations have been used in some artistic genres, like commensurazione. Art theory presents rules on how paintings should be executed, like using of hue and luster~\cite{Baxandall1985}. Also, analogies with other painting were utilized as well as analogies originated in scientific theories~\cite{Baxandall1985}. Artists use nature as inspiration, as they select and amplify certain aspects~\cite{Mondrian1995}. Insight is also important.  However, procedures in art also pursue a generative process~\cite{Baxandall1985,Mondrian1995}, in which existing images are deconstructed and reconstructed to produce a new meaning, like the equivalence between natural and spiritual~\cite{Mondrian1995}. Such a procedure is similar to hypothesis design and testing~\cite{Baxandall1985,Mondrian1995}, in which a new hypothesis is tested against previous artwork and the artist's rational (intentional) behavior~\cite{Baxandall1985}. Intentionality suggest the existence of causal relationships that use the spatial and temporal structure of a composition to support the story told by a painting~\cite{Baxandall1985}. This procedure creates not only a new artistic object, but also a new expression language, as explained by Shapiro: ``discriminating the good in an unfamiliar form that is often confused by the discouraging mass of insensitive imitations''~\cite{Schapiro1995}(pp. 16). The pursued constraints guarantee the novelty of the created art~\cite{Mondrian1945,Schapiro1995}. For example, Mondrian used constraints like using only vertical and horizontal lines, pure colors, and opposition of colors to remove vagueness and tragic, thus, offering a precise meanings to his paintings~\cite{Mondrian1945}. The selection of the features to be highlighted is arguably based on intuition~\cite{Mondrian1945} to achieve the artist's intentions related to broad societal goals and beliefs as well as artistic requirements, like balance, harmony, and order~\cite{Mondrian1995,Schapiro1995}. It can be argued that the trial-and-error steps created mutated descendants of previous artistic features~\cite{Baxandall1985,Schapiro1995,Simonton2010}, however, these descendants are only tokens of the new expression language that is created. An important characteristics of the language used in abstract art, including Mondrian's work, is its capability to describe ambiguity, so that understanding ambiguity is tractable (solvable). 


Depending on how the decomposed subproblems are tackled and integrated, there are two opposite approaches in problem solving, top-down and bottom-up procedures. Both consider a hierarchical description of concepts, ideas, and relations. Top-down procedures assume that there is a general plan that defines the main concepts and their relations. Details are progressively added until the design is completed. Top-down design is popular in circuit design~\cite{Gielen2000,Medeiro1999}. In painting, commensurazione assumes a three-step, top-down process: devising the general plan that includes profiles and contours, defining relations through proportions between contours, and devising the features, like detailing and coloring~\cite{Baxandall1985}. Another instance of top-down design is to define creating artwork as the process of assigning values to the variables of a visual template (e.g., position, dimension, color, and cues to engage the viewer, like expected visual scanning), so that the desired meaning is communicated~\cite{Mondrian1995}. Bottom-up procedures focus on devising first the detailed concepts, which are then gradually integrated into the overall solution. A broad range of procedural approaches can be imagined by combining top-down and bottom-up solving of the subproblems into which a problem is decomposed. 

Applying a procedure to solve a problem is connected to previous and future work, including influence of earlier paintings and other traditions~\cite{Baxandall1985}. The current application continues previous applications of the procedure, and is continued by future applications~\cite{Schapiro1995}. Therefore, it has been argued that an artist's work is a logical development of previous art, including the work of others~\cite{Mondrian1945,Schapiro1995}. However, the continuation also includes a distinction between two art objects~\cite{Baxandall1985,Mondrian1945}. Therefore, a procedure extend beyond solving the current problem, like creating a certain painting, to pursuing a broad idea (goal) of the artist~\cite{Baxandall1985}. The reinforcement (through repetition) of artistic features with a certain purpose can establish beliefs, like elimination of form in painting serving the purpose of cementing freedom of expression~\cite{Mondrian1945}(pp. 38). 

Finally, learning and getting insight while conducting the procedure is critical considering the evolutive nature of the process. Learning can be conscious or unconscious~\cite{Mondrian1945,Mondrian1995}. Mondrian argues that art is an evolution process during which the artist uses multiple points of view, discovers new ways to express certain ideas, and learns about them by comparing them with known features (including features from other domains, like architecture~\cite{Mondrian1995}) and further meditating (generalizing) about their expression power~\cite{Mondrian1995}. For example, the artist can learn about the effect amplifying or reducing illumination on the expressed meaning~\cite{Mondrian1995}. Other learned elements include new constraints, rules, and invariant relationships. A better understanding of the artist's intentions and goals might also result by observing the meanings obtained through the new features and concept associations~\cite{Baxandall1985}. More precisely, Mondrian argues the importance of ``looking deeply'' to ``perceive abstractly''~\cite{Mondrian1995}. This suggests a repeated analysis of an artwork during which any newly acquired knowledge is used for the next analysis iteration~\cite{Mondrian1945}. Learning can also refer to the execution of a physical painting, like materials, textures, detailing, finishing, and so on~\cite{Mondrian1995}.

\subsubsection{(6) Beliefs-(7) Expectations:}

Beliefs are invariant ideas, rules, or their characteristics. These invariants over a certain time period can apply to an individual, group, or an entire culture. Hence, the large variety of ideas, rules, and characteristics implies that there is a large variety of beliefs too. The proposed model argues that beliefs are necessary in real life to tackle the huge complexity of the semantic space, if all variables are free. Many variables are correlated, and therefore analyzing the entire space of possibilities for a new idea, rule, or routine might be difficult in a reasonable amount of time. Instead, beliefs lock some of the free variables through priorities, importance, specific meanings (including symbolism, like the religious meaning of water and doves), a certain way of solving a problem, and so on, hence significantly reducing the semantic space of possible meanings, and making problem solving more tractable. Due to their existence over longer periods of time, beliefs are expected to produce a certain cognitive development of the members, such as adhering to common ideas, and also certain priorities, habits and skills to be used in the future.   

As they are invariant over a period of time, beliefs are likely to produce a certain set of central objectives (problems, needs), a specific way of communication, a particular set of metrics through which an artwork is evaluated to achieve its set intentions, as well as distinct way in which the observers are expected to interpret an artwork. Moreover, beliefs that exist over longer time periods are likely to produce knowledge that is systematized into top-down sequences of activities for the creation of a new painting, like the painting process that has three steps, general planning, painting the profiles and contours, and coloring~\cite{Baxandall1985}. In this top-down problem-solving process, general planning decides the overall composition, the position, size, and relations between the main forms, followed by detailing the overall composition through precise profiles and contours. Finally, coloring completes the detailing of the composition. This top-down process requires less experimenting through trial-and-error to identify the best painting outcome, as many of the related variables, like an optimized way of communicating a message through a composition were already decided. In this scenario, arguably, the main solution in achieving the intention of the customer was finding a painter that possessed the required knowledge and skills, not so much an artist that would innovate. It might explain the importance of prestige and the transmitting of the craft through mentorship.     

Beliefs can be explicit, such as the meaning of religious episodes as explained by Scripture, or implicit, i.e. following certain actions of a community, even though the beliefs behind those actions might not be known. Beliefs can evolve over time, such through refinements and adaptations, as their future expressions do not conflict with the present forms. Alternatively, beliefs can be replaced over time by newer beliefs that contradict them. 

Beliefs can originate in mathematics, sciences, religion, philosophy, and social norms of a culture. For example, the invariant ideas and rules used in paintings of the romantic period are based in geometry of planes, proportions, and perspective. Similarly, beliefs used in the paintings of the Viennese school of the XIX century are arguably grounded in insight from psychology~\cite{Kandel2016}. Religion imposed beliefs not only about the precise meaning of the episodes in the Bible, but also about the fact that their representation through art must be visually precise, pious, and memorable~\cite{Baxandall1985}. Beliefs over the superiority of rationality over emotions lead to goals of eliminating the subjective in an attempt to show absolute beauty~\cite{Mondrian1995}. Social norms imposed beliefs, like the preference of religious and historical themes over landscape~\cite{Baxandall1985} and depicting in detail rich garments to suggest a status of prestige and power of the clients that ordered a certain artwork. Beliefs can also originate in constraints related to previous artwork, such as the goal of not painting motion~\cite{Mondrian1945} and the subsequent beliefs about the connection between natural, emotion and diminishing beauty, the limitations of natural representations, the importance of lines, colors, and integration of duality to achieve harmony and balance, the purpose of art, the evolution of life from natural to abstract, and the characteristics of the societies of the future~\cite{Mondrian1995}.           

\subsubsection{(8) Outcomes:}

Outcomes refers to completed circuit designs or finished paintings in the case of Mondrian's work.

\section {Case Study: Understanding Mondrian's Paintings Vs. Understanding Electronic Circuit Designs}

\subsection {Circuit Design}

{\bf Addressed problems}. Electronic circuit design usually implies solving ill-defined problems, for which solutions must tackle conflicting requirements, like satisfying or improving one requirement simultaneously worsens another requirement~\cite{Doboli2014}. Conflicting requirements originate performance tradeoffs in circuit design, and are an essential aspect of the design process~\cite{Gielen2000}. Typical tradeoffs between conflicting requirements are amplification Vs. bandwidth and stability, and speed Vs. low power consumption. The meaning (semantics) of the utilized design elements, e.g., BBs (including MOSFET transistors) and design rules, is defined to a large degree by laws of physics, even though there is some ambiguity due to electrical properties that were previously minor (e.g., effects due to shrinking sizes of MOSFET transistors) or unwanted poles and zeros when connecting subcircuits together~\cite{Tang2006}. Main economic constraints of design require producing solutions in the shortest amount of time, with the lowest costs, and with as few errors as possible. Utilizing novel design features is not justified unless it has been proven that current solutions cannot address the application needs. Therefore, reuse of previous design features is often pursued. Moreover, design solutions can be objectively compared with each other based on their numerical performance~\cite{Gielen2000}, leaving little room for subjective interpretations. 

\begin{figure}
\includegraphics[width=\textwidth]{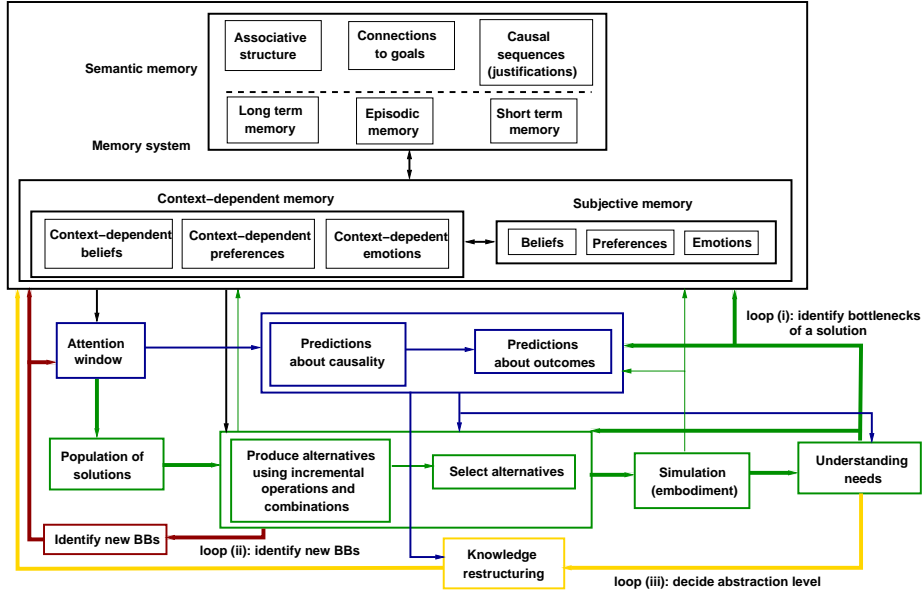}
\caption{The cognitive architecture InnovA for the design of new electronic circuits.} 
\label{cog_arch}
\end{figure}

Figure~\ref{cog_arch} illustrates the cognitive architecture (called InnovA) that was proposed for devising new electronic circuits by computationally mimicking the cognitive activities in problem solving~\cite{Li2018}. 

{\bf Memory}. The cognitive architecture has a memory system organized as three parts: First, the {\tt Memory system} includes the design knowledge available for problem solving. It is organized as {\tt Long-term memory} that includes all design knowledge, {\tt Short-term memory} that keeps the recently used knowledge, and {\tt Episodic memory} that stores the previous experience of using specific design knowledge in problem solving. Second, the {\tt Semantic memory} represents the meaning of the design knowledge stored in the memory system. It is represented using three structures: the {\tt Associative structure} groups the design elements into hierarchical sets based on the similarity of the elements in a set. Associations connect subcircuits with similar functions (synonyms) or similar structures (homonyms). {\tt Connections to goals} structure indicates the purpose of each design element, and {\tt Causal sequences} structure presents how the design elements are linked together to produce a design solution. Third, {\tt Subjective memory} stores beliefs and preferences about using specific design knowledge (e.g., subcircuits) in a design solution. They are important in assessing the importance and hence in ordering the way in which unwanted features of a solution are addressed, like noise and nonlinear behavior. {\tt Emotions} module mimics the purpose of human emotions during decision making, like controlling the switching between pursuing a global or a local view of the design problem. For example, addressing very precise design needs focuses the process on certain local parts of a design, similar to how local processing is achieved through negative emotions, like anxiety and frustration. In contrast, positive emotions, like easiness, encourage a global perspective on the design problem~\cite{Romero}. {\tt Context-dependent memory} module retains the elements of {\tt Subjective memory} that have been used for solving the current problem.   

In addition, the architecture has three subsystems: (i)~the attention and prediction subsystems to relate (compare) a new problem or solution to previous similar instances and to predict the impact of the differences both in terms of overall operation and outcomes, (ii)~the reasoning subsystem to produce an explained solution (e.g., a circuit structure) for a problem, and (iii)~the knowledge extension subsystem, which creates new subcircuits (BBs) for future problem solving, and restructures the knowledge to incorporate the new information that was acquired during design. The operation of the cognitive architecture realizes three feedback loops. The three subsystems and the associated loops are presented next. 

{\bf Attention and prediction subsystem}. It is the part highlighted in blue in Figure~\ref{cog_arch}. The process starts by having the {\tt Attention~window} module being activated by unexpected features of the design requirements or circuit design, which is being analyzed as a potential solution to a posed design problem. For example, the design requirements that supported the devising discussed in~\cite{Jiao2015b,Lopez2005} required creating a solution for low voltage, low power applications and sufficient amplification and speed, e.g., slew rate, requirements. Attention is focused on the conflicting requirements, such as low voltage Vs. low power, and amplification and speed Vs. low power. Addressing one of the requirements adds supplementary constraints on the opposing requirement. For example, a solution might use sub-structures (BBs) from two different circuits: adaptive biasing class AB input and a three-stage with frequency compensation to improve stability. Based on their previous usages, hence the knowledge about the causal connection of the sub-structures (module {\tt Predictions about causality}) to their outcomes (module {\tt Predictions about outcomes}), adaptive biasing class AB input with common-mode feedback is expected to offer high amplification and speed (e.g., slew rate). Similarly, the three-stage structure with frequency compensation is expected to improve the stability of the solution. Making predictions about causality and outcomes might involve the {\tt Context-dependent memory} module, which includes beliefs, priorities, and emotions specific to the application that is being addressed. As the current application might be similar to previous applications, {\tt Context-dependent memory} module is connected also to {\tt Subjective memory} module that stores the beliefs, priorities, and emotions previously acquired during design. 

{\bf Reasoning subsystem}. The part highlighted in green in Figure~\ref{cog_arch} indicates the reasoning part to create a solution that is verified to satisfy the requirements. The substructures selected through attention and prediction become part of {\tt Population of solutions}, which includes all the circuit elements that can be relevant in devising the solution. Module {\tt Produce alternatives using incremental operations and combinations} modifies the selected substructures to adjust them to the problem requirements, and/or combines them to build new designs (concept combinations). Incremental operations create local changes to a substructure without changing the main nature of the substructure. Module {\tt Select alternatives} selects among the alternatives available to solve a problem the alternative that is further considered at the current step. Module {\tt Simulation} produces a complete characterization of the design, hence fully describing its meaning (semantics) with respect to the problem requirements. Module {\tt Understanding needs} compares the characteristics of the current design with the requirements, and then identifies the causes for their mismatches, including the reasons for the unsatisfied requirements.  

{\bf Updating}. Module {\tt Identify new BBs} recognizes new substructures that are generated by adjusting present BBs. A BB is a subcircuit that can be used for solving a wider set of problems, hence is not customized only for the current application. Module {\tt Knowledge restructuring} changes the semantic memory because its current structuring does not lead to finding a solution that addresses the application requirements. 

The architecture memory and subsystems implement three nested loops as shown on the figure. The inner most loop, called loop~(i), aims at finding the bottlenecks of a solution. This is critical in ill-defined problem solving, as devising solutions that tackle opposing requirements is the main challenge of such problems. It repeatedly identifies unexpected or unusual features, which are then used through reasoning to understand the next design need to be addressed. This loop can lead to situations in which new subcircuits (BBs) are being created as a result of incremental changes of the existing design features because of the needs that must be tackled. A second loop emerges, called loop~(ii), because a new subcircuit draws attention and supports new predictions, which then subsequently can be used in adjustments by incremental operations and in combinations with other subcircuits. The outermost loop, called loop~(iii) in the figure, is executed if the two previous loops fail in addressing the conflicting requirements by using the available subcircuits or by creating new BBs. In this case, the design process must consider a higher abstraction level that would present the source that causes the unreconcilable performance requirements. The attention and prediction subsystem focuses on the cause and then reasons about solutions that would reduce the impact and nature of the causes on the overall solution characteristics. This loop produces knowledge restructuring.

\subsection {Mondrian's Paintings}

{\bf Addressed problems}. Mondrian offers a detailed presentation of the ideas behind his work, including the factors that originated the originality of his paintings~\cite{Mondrian1945,Mondrian1995}. For example, he stated in~\cite{Mondrian1945} that he ``disliked particular movement'' (pp. 10) and that he wanted to paint ``not bouquets, but a single flower a time'' (pp. 10). Mondrian belonged to the abstractionist movement in art, and was inspired by a number of art styles, like Impressionism, Fauvism, and Cubism~\cite{Mondrian1945}. However, his goal of creating original artwork led to the observation that Cubism, to which he initially participated, does not eliminate all triggers of subjective feelings, like natural forms, and hence, a logical extension would be to pursue representations that achieve pure, timeless beauty by presenting pure reality that is void of particulars~\cite{Mondrian1945,Mondrian1995}. The goal for art would be to describe human condition in a modern age, such as a mechanized age dominated by materialism~\cite{Mondrian1995}. His beliefs, ideas, and their expression in paintings evolved over time, as he explored new ideas and approaches.

\cite{Mondrian1945} argues that art styles follow a continuous evolution process towards a ``clearer content of art'' (page 17). The evolution process incorporates new ideas in science, philosophy, and society. Therefore, restructuring should reflect this evolution. It offers legitimacy to a new art style within a specific cultural frame~\cite{Baxandall1985}. Moreover, there is a consistency (coherence) of the artwork within a certain style with respect to agreeing with the beliefs and goals of other work of the style, as well as a consistency with the artists previous work~\cite{Baxandall1985}. Finally, there is a necessity element that justifies the need for restructuring and new beliefs and goals as requirements to express the new characteristics of a society~\cite{Baxandall1985}.      

\begin{figure}
\includegraphics[width=\textwidth]{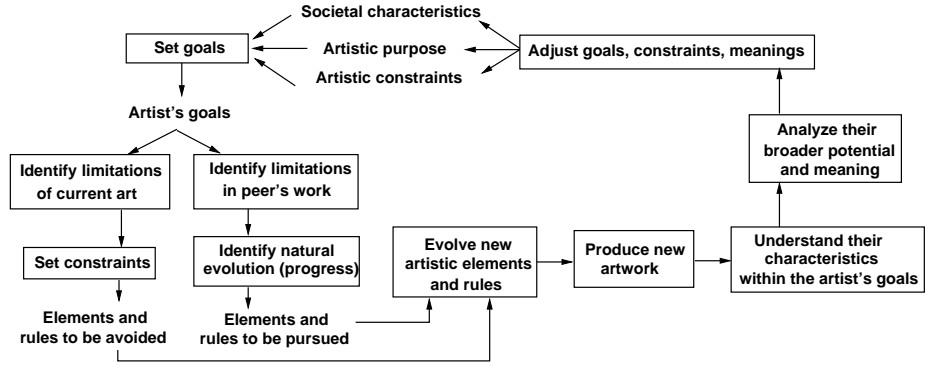}
\caption{Open-ended problem-solving process corresponding to Mondrian's painting.} 
\label{Mondrian_overall}
\end{figure}

{\bf Problem-solving process}. It can be argued that Mondrian's efforts to create new paintings is an open-ended problem-solving process that was based on a number of constraints on the ideas to be pursued and avoided, as well as his personal goal to produce paintings that are original and contribute to the mission of art as opposed to other domains like architecture and decorations. While his work was not rooted from the beginning in a precise set of rules and elements to be used in constructing new outcomes, it was guided by the elements that were not be utilized in the creative process, thus by a precise evaluation mechanism. The focus on distinguishing from previous work leads to evolving new elements that articulate this distinction, and which lead to the identification and exploration of a new solution space that can be only partially predicted by the previous work. Solving the open-ended problem posed by the goal of creating original art in-tune with the current society meant devising a new expression language (not only particular paintings), including the elements, rules, and objectives of the language. The uncovered solution space is formed by all paintings created using this expression language. The ambiguity of the BBs, ideas, and rules is broad, as Mondrian's work proposes a reinterpretation of the meaning (semantics) of lines, forms, and colors in a way that departs from their traditional meanings based on nature. The interpretation of his effort is subjective, even though newer work attempts to offer a more quantitative evaluation based on the neural processing of the brain~\cite{Onians2007,Zeki1993}. While the new meanings support expressions beyond visual elements, they raise significant challenges in terms of understanding the semantic ambiguity that emerges, so that new meanings are possible based on an observer's interpretation~\cite{Onians2007}. Hence an outcome does not have a deterministic meaning anymore, but is rather a guiding template to generate new interpretations by the viewers.

Figure~\ref{Mondrian_overall} depicts this process of solving open-ended problems. The process continuously uses two kinds of constraints: elements and rules to be avoided as they express limitations of previous artistic styles, and artistic elements and rules to be pursued as they reflect the natural evolution (progress) of the artistic style with which the artist identifies himself with. These two kinds of constraints support then the evolving of new artistic elements and rules that are used by the artist to produce new artwork. The artwork is evaluated to understand its characteristics within the artist's goals, and then analyzed to understand its broader meaning and potential to further support original work. The analysis can lead to an adjustment of the artist's goals, constraints, and meanings of the utilized concepts. Note that the process does not only generate individual art objects, but it produces a new language to express the solution space of the open-ended problem.  

\begin{figure}
\includegraphics[width=\textwidth]{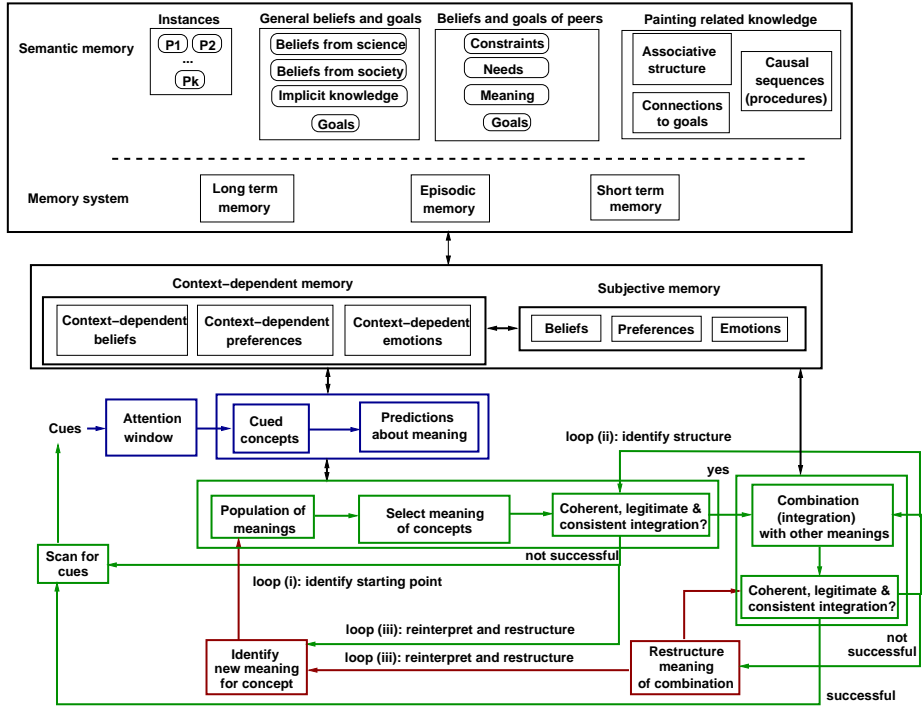}
\caption{Possible cognitive architecture for insight into Mondrian's paintings.} 
\label{cog_arch_M}
\end{figure}

Figure~\ref{cog_arch_M} depicts a possible cognitive architecture for getting insight into Mondrian's work. It corresponds to the problem solving flow in Figure~\ref{Mondrian_overall}. To easy the comparison with the architecture for circuit design, the presentation of the architecture was devised similarly to the architecture in Figure~\ref{cog_arch}. 

{\bf Memory}. The memory system of the cognitive architecture in Figure~\ref{cog_arch_M} has the same broad structure as that of the cognitive architecture for electronic design in Figure~\ref{cog_arch}, but the stored knowledge and knowledge organization is different, e.g., shown for the {\tt Semantic memory} module. Instead of a hierarchical structure organized based on similarities, causal connections to goals, and causal sequences to create a design solution like in Figure~\ref{cog_arch}, the semantic memory is based on {\tt Instances} of paintings, i.e. paintings {\tt P1}, {\tt P2}, ..., {\tt Pk}, which represent the artwork with unique and deciding influence on the artist. For example, Mondrian mentions Futurism, Dadaism, Surrealism, and Cubism as currents that influenced him (page 18 in~\cite{Mondrian1945}), so the module would include a sample of paintings of these artistic currents. However, these instances are not organized in a hierarchical structure along features (like in circuit design), but instead are kept as separate instances. Instances materialize the constraints, like features to be pursued and to be avoided, and hence, they can be encoded over time into generalizations and abstractions that can originate new beliefs and goals. Module {\tt General beliefs and goals} includes explicit ideas and facts largely accepted by a society, and which originate in science (e.g., quantum physics and the theory of relativity), philosophy and aesthetics (like dualism and harmony~\cite{Mondrian1995}), and society (i.e. the dual nature, natural and spiritual of men~\cite{Mondrian1995}). In addition to explicit beliefs, implicit beliefs are those accepted without a motivation, like trends in a society, like an increased emphasis on material aspects of life. Goals represent the broader goals set for society, such as the desire to build a happier, more intellectual society~\cite{Mondrian1995}. This module corresponds to {\tt Societal characteristics} in Figure~\ref{Mondrian_overall}. Module {\tt Beliefs and goals of peers} provides focused, art-related beliefs and goals of the art community to which the artist participated, like beliefs about the role and color in artistic expression or the purpose of art, such as its role in the formation and preservation of beauty~\cite{Mondrian1995}. Their purpose is to articulate the specifics of current art through constraints, like differences from previous art styles and means to emphasize these differences~\cite{Stokes2006}, needs, i.e. limitations that an art trend attempts to address as compared to previous art styles, goals set for the present style, and the meaning assigned to specific forms of representation, like line and color. They store the {\tt Artistic purpose} and {\tt Artistic constraints} in Figure~\ref{Mondrian_overall}. From a formal point of view, they are rules expressed using various logic systems.  

Modules {\tt Subjective memory} and {\tt Context-dependent memory} include the artist's subjective beliefs and goals, which are instantiated, adapted, and modified from the general beliefs and goals as well as those of his/her peers. For example, Mondrian's writings~\cite{Mondrian1945,Mondrian1995} mention a large set of goals, constraints, needs, meanings and goals, which were pursued in his artwork. Some of his beliefs and goals are as follows: constraints - avoid natural forms and colors as they produce subjective, tragic reactions, and deny normal perspective to force a new way of understanding art; needs - find a new way of expression to avoid the diminishing appreciation of natural beauty, use pure color (like black, white, red, blue, yellow), and devise abstract representations based on lines but no forms; meaning - establish the equivalence of beauty with harmonious, equilibrated duality, use straight lines to avoid the variability in nature, and utilize horizontal and vertical lines to express opposite relations; and goals - remove the individuality to produce universality, describe the immensity of nature by expressing expansion, rest and unity, and express the multiple facets of truth. The two modules store {\tt Elements and rules to be avoided} and {\tt Elements and rules to be pursued} in Figure~\ref{Mondrian_overall}.

{\bf Attention and prediction subsystem}. The subsystem is highlighted in blue in the figure. Cues are differences between the current image (e.g., a Mondrian painting) and instances in the semantic memory as well as beliefs in the subjective memory. Cues guide attention (through module {\tt Attention window}). Cues that draw attention pertain to the embedding of the architecture in the real world as well as to learned and observed features. Cues related to embedding mimic elements that are hardcoded in the brain~\cite{Zeki1993}, like cue centrality, size, colors, contrast and opposition, and globality Vs. detailing. Learned cues include situations that have been discussed in the literature, e.g., using unfinished lines in a composition~\cite{Deicher1999}, or interpreting a painting as an image observed through a window (analogy). Cues can also represent unexpected observations, like suddenly noticing different shades of white of neighboring surfaces. For example, the granularity of the painting focuses attention on the entire image, while crowded images focus attention on the details. Other cues are large and central elements, like having long, black lines or bright, red patches in the middle of an image. Cues can also be known (in the semantic memory) or previously seen, visual elements (in the subjective memory). 

As a result of cuing, concepts are retrieved from the semantic memory (module {\tt Cued concepts}) together with their associated meaning, also from the semantic memory (module {\tt Predictions about meaning}). For example, colors introduce a certain feeling, like black, purple and orange can produce a negative feeling of impersonality and lifelessness. Feelings induced by colors can also suggest movement, like the positioning of opposite colors on the diagonal, or 3D stacking, like the placing side by side of black and yellow. Similarly, paintings with few lines and a lot of white spaces create feelings of simplicity, order, and lightness, while crowded images of many colors produce feelings of difficulty and tension. The association of the predicted meanings to the cued concepts happens automatically as stored in module {\tt Painting related knowledge}, such as {\tt Associative structure} and {\tt Connections to goals}. The subsystem corresponds to activities {\tt Understand their characteristics within the artist's goals} and {\tt Analyze their broader potential and meaning} in Figure~\ref{Mondrian_overall}. 

Cues have an important role in understanding the meaning of an image, as they act as starting points in piecing together that meaning. A cue decomposes an image into components that can have an associated meaning. For example, vertical and horizontal black lines are used to separate yellow, red, or black surfaces, which create feelings, like heat, positivity, or movement in 3D. Lines do not act towards producing a composition made from forms, like in traditional painting, but as separators between elements with meanings. 

Each meaning of a painting element acts as a hypothesis that is further validated or modified as the remaining elements of a painting are understood. Predictions about meaning might incorporate laws of logic, geometry, and physics, i.e. the formation of shadows and how shadows relate to the positioning of layered surfaces. Also, using the continuation principle to explain the appearance of surfaces of color can lead to insight about the positioning of the surfaces, such as which surface is on top, and which is at the bottom. However, ambiguity of this meaning can emerge if the continuation principle is partially limited, so that surfaces are stripes. Analyzing the ambiguity can produce multiple meanings for the same image. Cues can force new interpretations, like instead of seeing the intersecting horizontal and vertical lines as a cross, they are used together with color to suggest the idea of movement~\cite{Deicher1999}. This subsystem corresponds to module {\tt Evolve new artistic elements and rules} in Figure~\ref{Mondrian_overall}. 

{\bf Reasoning subsystem}. The subsystem is highlighted in green, and it uses the meanings of the elements identified based on the cues to produce a meaning for the entire painting. {\tt Population of meanings} module stores the meanings that have been identified for the cued concepts, as explained in the previous paragraphs. A meaning to be further considered in reasoning is selected for integration ({\tt Select meaning of concepts} module). This meaning can be independently identified for the cued element, such as if a decision was made to sample different parts of the image, or it can be selected in conjunction with the meanings of the previously analyzed elements, such as the considered semantics includes horizontal, stacked layers of surfaces of color. The integration of new meanings into a previously hypothesized meaning serves to reinforce the correctness of the hypothesis. Integration must be coherent with the previously assumed meanings and the meaning of the cued concept (e.g., no contradictions), legitimate with respect to all stored beliefs, and consistent with previous semantic integrations (module {\tt Coherent, legitimate \& consistent integration?}). The subsystem corresponds to {\tt Evolve new artistic elements and rules}, {\tt Understand their characteristics within the artist's goals}, and {\tt Analyze their broader potential and meaning} in Figure~\ref{Mondrian_overall}.

The cognitive architecture implements three feedback loops. The first loop attempts to find the starting points for deciphering the meaning of a painting. If the meaning selection for a concept is not successful, the architecture scans for new cues (module {\tt Scan for cues}), which can lead to different semantic interpretations. The scanning process can jump to the next dominant cue or follow a systematic scanning of the entire image, such as from left to right and top to bottom. The second feedback loop searches for the structure that integrates the meanings of the cued elements. If the integration of the individual meanings is unsuccessful, the process considers different integrations (combinations) or different meanings for the concepts. The third loop reinterprets the meanings of concepts and restructures their integration, if the first two loops failed or if additional meanings were to be found.   

\begin{figure}
\includegraphics[width=\textwidth]{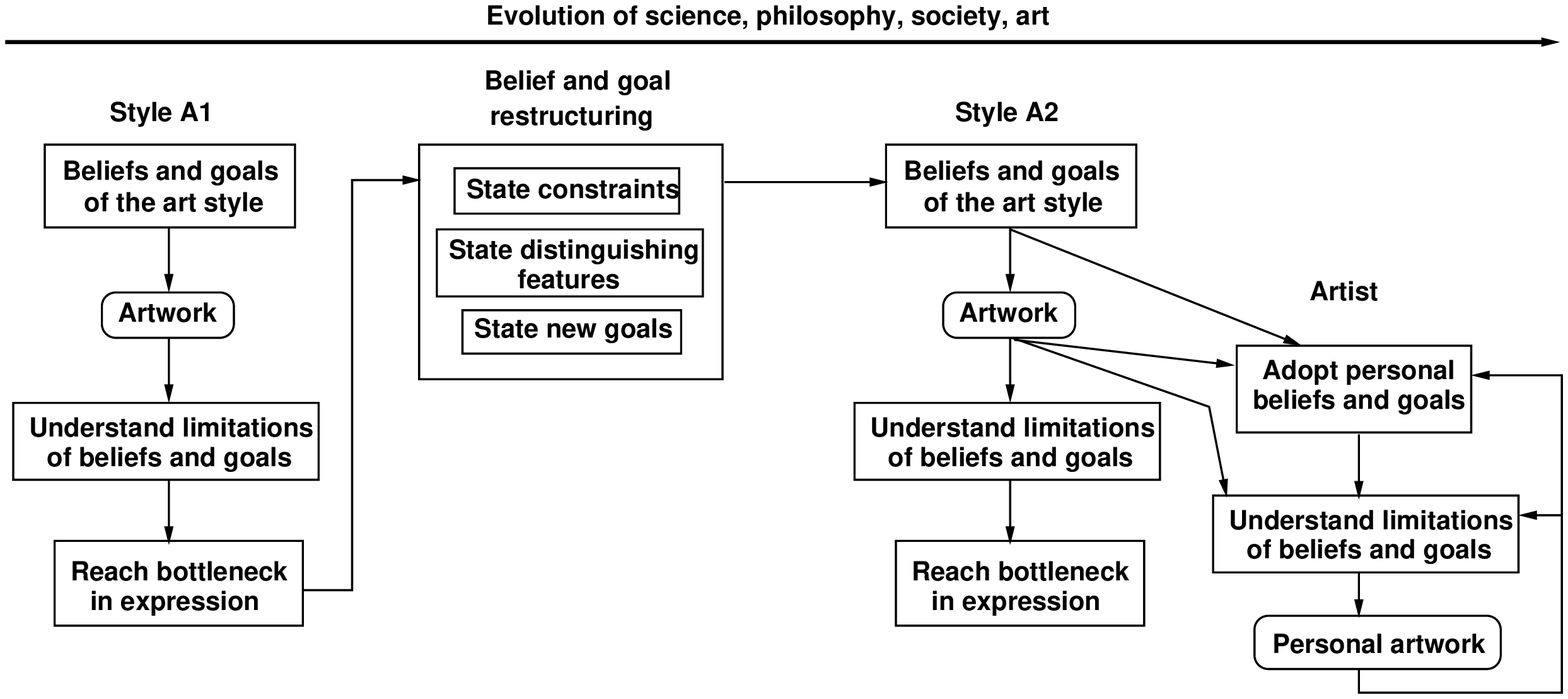}
\caption{Updating the semantic memory.} 
\label{cog_arch_M_details}
\end{figure}
               
{\bf Updating}. Figure~\ref{cog_arch_M_details} summarizes the process of updating the semantic memory over time. It refers to two generic art styles, called $A1$ and $A2$ in the figure. Artwork is created along the beliefs and goals set for style~$A1$. By evaluating the expression capabilities of the produced artwork, the limitations of style~$A1$ are understood, including the way these limitations origin in the beliefs and goals set for the style. Eventually, a bottleneck for style~$A1$ is reached, when it cannot further evolve to create new, original artwork beyond that already created. Belief and goal restructuring follows after reaching the bottleneck starting from the identified limitations of the style~$A1$~\cite{Mondrian1945}. Restructuring includes identifying constraints on what should and what should be not pursued by a new art style, stating of distinguishing features that supports differentiating it from previous styles, and stating of new goals. The new art style, called style~$A2$, develops its own beliefs and goals that are in sync with the restructuring elements. Within this new art style, an artist adopts his/her own personal beliefs and goals, which match to some degree the beliefs and goals of style~$A1$, but might also include different beliefs and goals based on the subjective interpretation of the artwork within the style. New personal artwork is created to reflect the personal beliefs and goals, which through analysis and evaluation leads to further adaptation (evolution) of the personal beliefs and goals, as well as new insight about the understanding, including limitations, of another artwork. This process corresponds to activities {\tt Identify limitations of current art}, {\tt Set constraints}, {\tt Identify limitations in peer's work}, {\tt Identify natural evolution (progress)}, and {\tt Adjust goals, constraints, meanings} in Figure~\ref{Mondrian_overall}.

\section {Discussion}

This section summarizes the main differences between problem-solving in electronic circuit design and understanding Mondrian's paintings as a starting point in identifying the computational methods that could attempt to characterize Mondrian's work similar to a human expert. Previous work showed that current Deep Neural Networks, like Convolutional Neural Networks, cannot classify well abstract artwork~\cite{Mahan2022}. There is a large body of work on computational methods for automated circuit design~\cite{Gielen2000}, but there is significantly less research on algorithmic approaches to characterizing art. The goal is to identify the new features required to process artwork as compared to other Computer Aided Design (CAD) activities.  

{\bf Addressed problems}. There are fundamental differences between the nature of ill-defined problems, like in electronic circuit design, and open-ended problems, such as creating new artwork. Ill-defined problems pose conflicting requirements and constraints, but which are usually known, such as expressed through numerical thresholds or ranges. Instead, open-ended problems impose a significant departure from the current solution space, as creating novel solutions is a main goal. There are usually no numerical descriptions for this departure. Also, there are no requirements for reducing cost and minimizing errors, like in engineering design. 

Due to their numerical descriptions, ill-defined problems in engineering can be objectively evaluated through mathematical methods based on physical models and precisely defined metrics, such as by using numerical simulators. They support introducing precise quality metrics, procedures to compare solutions, and to build surfaces that reflect the nature of tradeoffs between conflicting requirements, like Pareto surfaces. There are no similar approaches for open problem solving in art. There are currently no equivalent methods based on theories in science, which could lead to mathematical evaluations, even though recent work in neuroscience and neuroaesthetics suggests that such an effort might be possible to some degree~\cite{Kandel2016}\cite{Zeki1993}. Evaluation of artwork, including its meaning, degree of innovation, and comparison between individual paintings, is performed by experts.        

{\bf Problem solving process}. Solving ill-defined problems in engineering involves mainly searching for the desired features in the space of previous solutions, and if searching fails then creating (generating) new solutions to address the issue that produced the failure. Each solution has a well-defined meaning and purpose, like satisfying precise requirements. The process uses three steps: (i)~searching among previous solutions to find the features that are likely to address the problem requirements by finding the best balance between competing requirements, and then integrating these features, (ii)~if step~(i) fails, using analogies (which are abstractions) previously used in similar problems, e.g., problems that posed the same kind of constraints even though their numerical values were different, and (iii)~if steps~(i) and~(ii) fail, generating new solutions for specific needs guided by the limitations observed in the current solutions. Hence, the solving process does not have to generate novel solutions unless they are required, such as after the current solution space was exhaustively explored to lead to the conclusion that it cannot tackle a specific need. The nature of problem solving as well as engineering constraints, i.e. requirements rooted in day-to-day operation, minimizing cost and number of errors, emphasize reusing previous work and solutions, hence impose a Bayesian memory character to the problem solving process. 

In contrast, solving open-ended problems in Mondrian's artwork is less of a search among previous solutions, but mainly a generative process that produced a new way of visual expression, and thus uncovered an entire solution space represented by the paintings of this space. Each painting has an ambiguous meanings and purpose for an observer, even though the artist had well purpose when creating the work under his attempt to meet the overall goals of art. Hence, a painting acts as a template that guides the association of new meanings by an observer. The generative process is based on the need to be novel and unique, thus, to be distinguishable from previous artwork. This is achieved through a set of well-defined beliefs, goals, and examples of elements and features that must be avoided. The problem-solving process evolves a sequence of features that are based on previous work by the author while still meeting the avoidance criteria, and are legitimate, consistent, and necessary. The process has less of a Bayesian memory character, but more of a restructuring (redefinition) of the expression mechanism used in creating art.    

Note that work on computational methods in circuit design has proposed using evolutionary mechanism, like Genetic Algorithms, to devise new circuits~\cite{Gielen2000}. However, these evolutionary processes do not contemplate the legitimacy, consistency, and necessity of each incremental modification step. These solutions have a low level of creativity~\cite{Ferent2011} and have not been adopted by circuit designers, even though they meet their requirements.            

{\bf Memory}. As discussed, the memory system for circuit design stores associations between design elements, connections between design elements and their roles, like the purpose they serve, and causal sequences describing procedures to solve a design problem. Associations indicate design elements with similar functions in a design and similar structures. Connections to roles serve as precise cues to access the memory depending on the specific design needs. The structure and meaning of each element are precisely defined by numerical values, like the values of physical concepts (i.e. charge, current, voltage), and metrics. Hence, each element is unique, and the similarity of their structure and functions also represents an approximation. Design elements are connected with each other through theories and laws in mathematics, science, and previous design experiences. The preference for reusing existing design elements supports understanding their capabilities and limitations. Abstractions in hierarchical knowledge representations are created, in which the abstractions describe general principles, which are instantiated by distinct design elements with different capabilities and limitations. The hierarchical structure supports design through high-level reasoning. Hence, the nature of memory organization supports a design process that evolves from bottom-up design during the initial stages to top-down design during the latter stages, as the hierarchical knowledge structure is created. 

The memory system for understanding Mondrian's paintings includes specific instances of paintings and painting features, but they are not grouped in hierarchical structures. Each instance is unique, and representative for a certain objective or subjective purpose. There can be, however, a clustering of the paintings based on their topic, structure (composition), or specific features (i.e. color). Separately, the memory system stores beliefs and goals from mathematics, science, and society as well as constraints, needs, meanings, and goals of peers and the artist. The last memory component changes through the completed artwork over time, some of the changes being explicitly articulated as a result of new insight while others remain implicit. There is a causality link between certain artistic features and subjective attention, feelings, and emotions as well as broader about the purpose of art, like pleasure and prestige. The connections to attention serve as cues and the connections to emotions and feelings to activating the broad meanings used in understanding a painting. However, the causality link is less based on explanations. The memory system has fewer solving procedures describing a causal sequence to create a new painting.  

The three memory subsystems are not related to each other through laws of mathematics and science (like in circuit design) but through analysis and discussions by experts. However, the connections between the three subsystems might not be fully explainable. The understanding of an artwork is consistent with the main beliefs and goals of the artist, but there can be inconsistencies with other beliefs and goals, including previous features that the artist stopped using, such as curved lines, which Mondrian stopped using in his later work. 

Artistic features and paintings can be ambiguous as there is no numerical definition grounded in the theories and laws of mathematics and science. The meaning of a feature can be extended or redefined by using it in a new context or for a new purpose. The degree to which the new meaning is valid depends on social aspects, like its acceptance by peers and public as well as its reusing to create new paintings.   

{\bf Attention and prediction}. In circuit design, attention goes to associating the design features, such as BBs, to the performance requirements. BBs are identified based on their structural similarity (e.g., form) based on BBs that are agreed on by the design community, hence vetted solutions. The prediction part assigns meaning to the identified BBs by causally connecting the BBs to the expected performance. Attention is also drawn to changes of the BBs from their previous structures, such as adding MOSFET transistors to the BB or sharing MOSFETs between different BBs. Assigning meaning to the changes by causally associating them to the changes in performance as demanded by requirements. Meaning assignment considering the information learned about capabilities and limitations from previous designs. Considering that new solutions reuse or adapt previous design features, the BB meaning has a Bayesian memory character as previous roles (purposes) in design are critical in understanding future meanings too. Therefore, understanding a design can be done only in the context of the design knowledge of previous solutions. The meaning of BBs is deterministic (i.e. with minimal ambiguity) as it is based on laws and theories in mathematics and sciences. Moreover, a complete meaning for all conditions and situations can be produced. This meaning does not change over time. 

Attention for Mondrian's work is drawn by visual cues, like color, size, centrality, contrasts, and unusualness. Recognizing these cues does not require any specific training, even though art training makes cue identification more effective and robust (e.g., certain cues are not missed). Cues can be of different kinds, including local and global cues, like the granularity of the grid in Mondrian's work. Cues also induce certain emotions and feelings to the observer. Thus, they can induce the overall template used to understand the meaning of a painting, like joy, energy, sadness, or motion. As identifying cues is important in understanding the meaning of a painting, the procedure for cue searching is critical, such as sequential search or search after repetitive patterns. 

Cues start the process of decomposing a painting into its composing elements or forcing a new perspective by forcing the viewer outside his / her comfort zone, such as contemplating another understanding than previous ones. Hence, cues might serve to annul the Bayesian memory character. Meaning can also change over time based on new beliefs and goals. 

The meaning of the elements composing a painting is ambiguous and depends on subjective interpretation. During the understanding process, hypotheses about their meaning are formulated and validated during the further understanding of a painting. Multiple meanings can result. The composition of the element meanings can follow a hypothesis testing process, in which the main hypothesis of what the composition means integrates the meanings associated to the composing elements, or a process that integrates bottom-up the element meanings, which are separately identified. The meaning of a painting acts as a semantic template that can accommodate different interpretations by distinct observers.  

{\bf Reasoning}. In circuit design, reasoning creates and verifies a circuit solution by explaining its operation. Reasoning is mostly performed locally through incremental modifications and combination of existing sub-structures and BBs to solve the requirements not met by the current design. The overall solution structure within which changes are made stays the same for most situations. The incremental steps are selected from a set of alternatives from previously devised solutions. Causal reasoning justifies a solution by indicating how a design feature is needed to accommodate the problem requirements. Restructuring the global structure of a design solution is justified only after the solution space of an existing structures are completely analyzed, hence changes are driven by understanding the limitations of the existing designs. The repetitive nature of the process can support the replacement over time of the initial, bottom-up design process with a more top-down process.

In art, the reasoning process is guided by the cues on which the observer's attention focuses. The cued concepts have an associated meaning that is used in reasoning and can also suggest a global meaning into which the meaning of other identified concepts is integrated. Reasoning implements hypothesis testing but without having an evaluation method based on mathematical or scientific theories and methods or numerical evaluations using metrics. Instead, the correctness of meaning integration is based on criteria, like legitimacy, consistency, and necessity, which are reinforced by previous, successful integrations.   

{\bf Update}. In circuit design, new knowledge about BB and substructure meanings and design procedures is learned by comparing the requirements of the current problem with the requirements of previously solved problems. Knowledge updates are justified by their causal connections with their roles in circuit design. The capabilities and limitations of BBs, substructures, and design procedures are updated as new designs are completed. New BBs and other circuit substructures are also devised and learned. Further understanding of the design challenges of ill-defined problems results by relating the differences in requirements and the specific designs to address the differences. Over time, solving similar problems supports the understanding of the effectiveness of the constructed solutions. 

In art, previous using of artistic ideas and features explores the capabilities and limitations of the artist's approach and the style of the paintings, and then connects the capabilities and limitations to beliefs and goals. Knowledge update also includes the constraints about ideas and features that should not be pursued by the artist, hence supporting the identification over time of elements that should be pursued, as part of the artist's goals, beliefs, and expressive language. Note that these constraints are very different in nature than constraints in circuits design, as they do not indicate that the entire solution space eliminated through constraints was already explored.    

\subsection {ML Approach to Identify Mondrian's Work in a Set of Paintings}

Sections 3.1 and 4.2 give a comprehensive description of the characteristics (e.g., ontology) of the eight components and the computational flow of a cognitive architecture that uses the components to understand artwork, such as Mondrian's paintings. This subsection explains how the components and flow can be utilized as a starting point to devise new computational processing, such as distinguishing artwork that is mainly based on non-exhibited properties (NEXP), like in modern art. As explained in~\cite{Mahan2022}, NEXPs correlate less to visual features, hence are hard to process using traditional DNNs. Hence, it is important to devise accurate and robust processing flows that can exploit the available data while minimizing the impact of missing information.   

Figure~\ref{cog_arch_M} illustrates a comprehensive cognitive architecture that we proposed to analyze and understand Mondrian's abstract paintings. However, implementing the architecture is challenging because of the implications on the architecture operation due to context dependency (e.g., culture, peers, and personal experience), emotions, and implicit processing. These elements influence knowledge organization and recall from the memory, including general beliefs and goals, perceived needs, meaning and goals by peers as well as identified constraints, and the agent's subjective associative connections, connection to goals and causal sequences. Subsequently, they impact the operation of the attention and prediction subsystem, reasoning subsystem, and updating of the architecture. These dependencies suggest the difficulty to devise an ML-based approach to understand and classify abstract painting, like Mondrian's work. 

\begin{figure}
\begin{center}
\includegraphics[width=0.8 \textwidth]{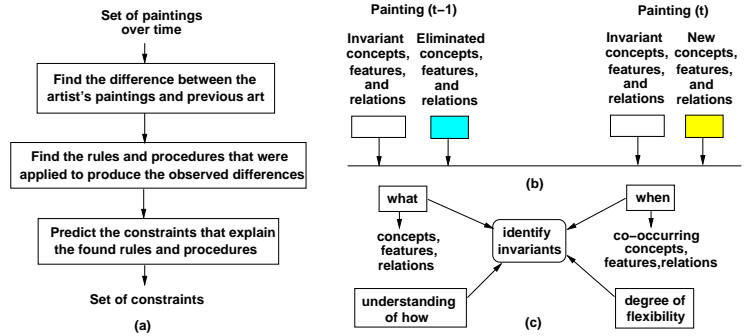}
\caption{Proposed ML flow to identify Mondrian's paintings.} 
\label{proposed_flow}
\end{center}
\end{figure}

This subsection presents a possible methodology that attempts to address the problem of distinguishing Mondrian's work from other paintings by incorporating activities that can be reliably performed with arguably less information about context, emotions, and implicit processing. It relies on the observation that artists, including Mondrian, identified explicit constraints that distinguished them from previous work and guided their work, including the considered topics, overall concepts, and implementations~\cite{Mondrian1945,Stokes2006}. They are due to the different beliefs and preferences of the artist. These constraints produced observable features of paintings, which can be found by comparing them with previous work. Following the reverse path from the output of the cognitive architecture in Figure~\ref{cog_arch_M}, the second step uses the observed differences to identify the rules and procedures that were likely used by the artist to create the observed differences. Finally, the third step identifies any invariants and patterns on how the artist used rules and procedures over time relate to the constraints that Mondrian used to distinguish his work from previous paintings. Among the characteristics in Section~3.1, the three steps are based only on those that can be extracted from visual features. Figure~\ref{proposed_flow}(a) illustrates the three steps.        

The first step should identify the differences between a painting by Mondrian and previous paintings by other artists and by Mondrian himself. The identified differences pertain to the category of concepts and ideas, discussed in Subsection~3.1. Without using information about context, emotions, and implicit knowledge, the concept characteristics that can be identified based on visual features relates mainly to concept description, like features and connection between features. It is difficult to precisely identify the concept meaning and the related meaning understanding process. However, some insight about concept meaning can be found, without understanding what that meaning is, since meaning depends on the used visual features and the other co-occurring concepts. Hence, partial information on concept similarity and integration with other concepts can be found. The specific metrics that can be computed include the following: feature similarity and differences with previously used concepts, new features, new concepts, frequency of concept occurrence in previous paintings, frequency of concept co-occurrences, new co-occurrences, and previous co-occurrences that were dropped in future work. The characteristics of ideas enumerated in Section~3.1 are partially covered by the metrics on groups of concepts, even though it is hard to infer more detailed insight on idea purpose, meaning, origin, characteristics, grouping, and evolution.        

The second step should find the rules and procedures that were likely used by Mondrian to produce the observed differences. A set of rules used to create a new painting, e.g., $Painting(t)$ in Figure~\ref{proposed_flow}(b), is expressed by the new concepts and features added to the painting as compared to previous paintings by Mondrian, i.e. $Painting(t-1)$, and the concepts and features that were eliminated in the new painting conditioned by the concepts and features that kept being used, hence remained invariant. The relations between new and previous concepts and features are also captured. From the rule characteristics enumerated in Section~3.1, the visual painting features utilized over time can be used to monitor when a rule is selected and used, the used concepts and features, presence of new concepts and features, inclusion of unexpected cues that could guide the viewer's attention, its degree of relatedness to other rules, its changes over time, its flexibility, and if it is global or local, deterministic or stochastic, flexible or invariant. Other characteristics about rules and procedures are hard to extract, like cuing, connection to goals and beliefs, achieved decomposition and aggregation, produced effects like associations, comparisons, generalizations, etc., purpose, i.e. inquiry, hypothesis formulation, analysis, explanation, etc., addressing of ambiguity, reinterpretation, rule origin and gradient, degree of subjectivity, and dependency on context. 

The third step identifies the invariants of the rules and procedures applied by the artist for his sequence of paintings. Figure~\ref{proposed_flow}(c) illustrates the step. Each rule identified in the previous step is characterized by the four shown components: component~{\em what} refers to the concepts, features and relations that were added and eliminated in a new painting as compared to the previous, component~{\em when} describes the unchanged concepts, features, and relations that co-occur with the new ones, component~{\em understanding of how} presents the sequence of individual changes that are the difference of a new painting from the previous paintings, and component {\em degree flexibility} expresses the amount of change between the sequence of individual changes for the current painting and the sequences of individual changes for the most similar paintings. Invariants are elements that tend to remain constant for the four components or the co-occurrence of elements from the four categories.

\section {Conclusions}

This paper aims to discuss and identify missing capabilities of popular parameterized computational models in Machine Learning, like Deep Neural Networks (DNNs), so that their semantic processing capabilities can possibly address activities beyond traditional classification tasks. Our previous work showed that existing DNNs cannot tackle well classification using semantic information (like abstractions and ambiguities) that is not a linear combination of visual features. The discussed work identifies the missing features by comparing the process of understanding artwork with the process of understanding electronic circuit design. Like art, circuit design is also a creative problem-solving activity, and for which our previous work proposed a cognitive architecture, a computational structure that loosely mimics human problem solving. The comparison methodology considers two semantic layers. The first layer tackles eight components, which are discussed in detail: goals, concepts, ideas, rules, procedures, beliefs, expectations, and outcomes. These elements are part of the cognitive process during problem solving and can be tied to the parts of a cognitive architecture for that activity, like memory, concept learning, representation and combination, affect, insight, and so on. The second layer describes a cognitive architecture for problem solving. It incorporates five elements: the nature of the problem, knowledge representation in the memory, the attention and prediction subsystem, the reasoning subsystem, and knowledge updating. The methodology was used to devise a computational method that can separate Mondrian's paintings from other paintings. Future work will further investigate the discussed ideas.

%
%
%
%

\end{document}